\documentclass{article}

\usepackage{microtype}
\usepackage{graphicx}
\usepackage{subfigure}
\usepackage{booktabs} % for professional tables

\usepackage{hyperref}

\usepackage[accepted]{icml2025}

\usepackage{amsmath}
\usepackage{amssymb}
\usepackage{mathtools}
\usepackage{amsthm}

\usepackage[capitalize,noabbrev]{cleveref}

\usepackage{tabularx}    % 自适应宽度表格
\usepackage{booktabs}    % 专业表格线
\usepackage{multirow}    % 跨行单元格
\usepackage{multicol}
\usepackage{array}       % 列格式控制
\usepackage{siunitx}     % 对齐±符号（可选）

\theoremstyle{plain}

\theoremstyle{definition}

\theoremstyle{remark}

\usepackage[textsize=tiny]{todonotes}

\icmltitlerunning{Preprint}

\begin{document}

\twocolumn[
\icmltitle{Disentanglement in Difference: Directly Learning Semantically Disentangled Representations by Maximizing Inter-Factor Differences}

% It is OKAY to include author information, even for blind
% submissions: the style file will automatically remove it for you
% unless you've provided the [accepted] option to the icml2025
% package.

% List of affiliations: The first argument should be a (short)
% identifier you will use later to specify author affiliations
% Academic affiliations should list Department, University, City, Region, Country
% Industry affiliations should list Company, City, Region, Country

% You can specify symbols, otherwise they are numbered in order.
% Ideally, you should not use this facility. Affiliations will be numbered
% in order of appearance and this is the preferred way.
\icmlsetsymbol{equal}{*}

\begin{icmlauthorlist}
\icmlauthor{Xingshen Zhang}{UJN}
\icmlauthor{Shuangrong Liu}{UJN}
\icmlauthor{Xintao Lu}{UJN}
\icmlauthor{Chaoran Pang}{UJN,Quancheng}
\icmlauthor{Lin Wang}{UJN,Quancheng}
\icmlauthor{Bo Yang}{Quancheng,UJN}
% \icmlauthor{Firstname7 Lastname7}{comp}
% %\icmlauthor{}{sch}
% \icmlauthor{Firstname8 Lastname8}{sch}
% \icmlauthor{Firstname8 Lastname8}{yyy,comp}
%\icmlauthor{}{sch}
%\icmlauthor{}{sch}
\end{icmlauthorlist}

\icmlaffiliation{UJN}{Shandong Key Laboratory of Ubiquitous Intelligent Computing, University of Jinan, Jinan 250022, China}
% \icmlaffiliation{Lab}{Quan Cheng Laboratory, Jinan,China}
\icmlaffiliation{Quancheng}{Quan Cheng Laboratory, Jinan 250100, China}

\icmlcorrespondingauthor{Lin Wang}{wangplanet@gmail.com}
\icmlcorrespondingauthor{Bo Yang}{yangbo@ujn.edu.cn}

% You may provide any keywords that you
% find helpful for describing your paper; these are used to populate
% the "keywords" metadata in the PDF but will not be shown in the document
\icmlkeywords{Machine Learning, ICML}

\vskip 0.3in
]

% this must go after the closing bracket ] following \twocolumn[ ...

% This command actually creates the footnote in the first column
% listing the affiliations and the copyright notice.
% The command takes one argument, which is text to display at the start of the footnote.
% The \icmlEqualContribution command is standard text for equal contribution.
% Remove it (just {}) if you do not need this facility.

\printAffiliationsAndNotice{}  
% leave blank if no need to mention equal contribution
% \printAffiliationsAndNotice{\icmlEqualContribution} % otherwise use the standard text.

\begin{abstract}
In this study, Disentanglement in Difference (DiD) is proposed to address the inherent inconsistency between the statistical independence of latent variables and the goal of semantic disentanglement in disentanglement representation learning. 
Conventional disentanglement methods achieve disentanglement representation by improving statistical independence among latent variables. However, the statistical independence of latent variables does not necessarily imply that they are semantically unrelated, thus, improving statistical independence does not always enhance disentanglement performance. To address the above issue, DiD is proposed to  directly learn semantic differences rather than the statistical independence of latent variables. 
In the DiD, a Difference Encoder is designed to measure the semantic differences; a contrastive loss function is established to facilitate inter-dimensional comparison. Both of them allow the model to directly differentiate and disentangle distinct semantic factors, thereby resolving the inconsistency between statistical independence and semantic disentanglement.
Experimental results on the dSprites and 3DShapes datasets demonstrate that the proposed DiD  outperforms existing mainstream methods across various disentanglement metrics.
\end{abstract}

\section{Introduction}
\label{Introduction}

Disentangled representation learning has been recognized as crucial for artificial intelligence to achieve genuine understanding of the world \cite{bengio2013representation,lake2017building}. Its core objective lies in acquiring the information-rich latent representation to describe the semantic factors or attributes that consist of the object\cite{higgins2018towards}. In the disentangled representation learning, independent properties of an object, such as color, size, shape, and pose, would be distilled and assigned to different latent dimensions without interference. This separation enables precise manipulation and interpretation of specific attributes. Disentangled representations are considered to hold significant potential for controllable image generation\cite{zhu2018visual,gabbay2019demystifying,gabbay2021scaling}, reducing sample complexity \cite{bengio2013representation,ridgeway2018learning,van2019disentangled}, interpretability \cite{bengio2013representation,higgins2017beta}, and performance enhancement in downstream tasks \cite{locatello2019challenging,locatello2019disentangling,bottou2007scalling,higgins2018towards,peters2017elements,schmidhuber1996semilinear,tschannen2018recent}.

Many existing disentanglement representation learning approaches indirectly encourage disentanglement by minimizing statistical dependencies between latent variables (e.g., Total Correlation(TC)\cite{watanabe1960information}). The underlying assumption is that statistical independence may correlate with semantic factor disentanglement. Consequently, if latent variables are statistically independent, each variable is presumed to be more likely to encode a single independent factor from the data generation process. For example, $\beta$-VAE \cite{higgins2017beta} enforces alignment of the latent variable distribution $q_{\phi}({z})$ with a Gaussian prior $p({z})$ by increasing the weight $\beta$ of the KL divergence term in the variational lower bound. Although this method does not explicitly penalize statistical dependencies between latent dimensions, the increased $\beta$ value indirectly encourages alignment with independent priors, thereby partially reducing inter-dimensional correlations. FactorVAE \cite{kim2018disentangling} and $\beta$-TCVAE explicitly introduce Total Correlation (TC) as a regularization term, directly measuring and penalizing the KL divergence between the joint distribution $q_{\phi}({z})$ and the product of marginal distributions $\prod_j q_{\phi}(z_j)$. Minimizing the TC term effectively reduces statistical dependencies among latent variables. DIP-VAE \cite{kumar2017variational} proposes to diminish linear dependencies between dimensions by constraining the covariance matrix of latent variables to align with the identity matrix.

However, the computation of statistical dependencies faces inherent limitations. As noted by \cite{locatello2019challenging}, methods relying on posterior distribution alignment with standard normal distributions through sampling strategies can only reduce sample-level statistical dependencies while failing to ensure dimension-level independence. Moreover, penalizing statistical dependencies does not equate to improving semantic independence or enhancing model disentanglement capability. Empirical analyses reveal that as total correlation between latent variables decreases, disentanglement metrics do not exhibit consistent improvement -- statistical independence gains do not directly translate to semantic disentanglement progress.

To address these limitations, our method aims to directly learn semantic distinctions between factors. Intuitively, for samples generated by variations of the same latent factor (e.g., images of the same object with different colors or the same person with varying expressions), their latent representations should form compact clusters in the latent space. Conversely, samples generated by different latent factors (e.g., images with color variations versus shape variations) should maintain significant separation in the latent space. This implies that variations from the same factor should exhibit local consistency in the latent space, while variations from different factors should create substantial separation. The fundamental hypothesis is that in an ideally disentangled latent factor set, the variation magnitude in latent space caused by the same factor should be significantly smaller than that induced by different factors.

Inspired by this perspective, we propose \textbf{Disentanglement in Difference (DiD)} -- a novel framework that directly enhances disentanglement by explicitly maximizing the distance between latent representations affected by different factors. This strategy aligns more closely with the intuitive definition of disentanglement, where distinct controlling factors should maintain distinguishable representations. Our main contributions are fourfold:

\begin{itemize}
    \item A paradigm shift from indirect statistical independence constraints to direct learning of semantic factor differences through inter-sample variations
    \item DiD architecture design featuring a contrastive difference encoder that learns semantic variations across dimensions while enforcing inter-dimensional contrastive constraints
    \item Development of a contrastive-based loss function that explicitly amplifies distinctions between different factors
    \item Experimental validation showing state-of-the-art disentanglement performance on benchmark datasets (\textit{dSprites}, \textit{3DShapes})
\end{itemize}

\section{Related Work}

Current approaches in unsupervised disentangled representation learning are predominantly based on Variational Autoencoder (VAE) \cite{kingma2013auto} or information-theoretic Generative Adversarial Network (InfoGAN) \cite{chen2016infogan} frameworks. These methods generally share a core principle: introducing additional regularization terms into the model's loss function to reduce statistical dependencies among latent variables, thereby promoting disentangled representations.
\paragraph{VAE-based Methods.} Among VAE-based approaches, $\beta$-VAE \cite{higgins2017beta,burgess2018understanding} stands as a seminal work. By introducing a tunable hyperparameter $\beta$ into the Evidence Lower Bound (ELBO) loss function, this method constrains the posterior distribution of latent variables. While its objective is to encourage alignment between the posterior distribution and a predefined independent prior (typically an isotropic Gaussian), increasing the weight of the KL divergence term ($\beta > 1$) effectively restricts the latent space capacity and induces lower statistical dependencies across dimensions. However, selecting an optimal $\beta$ remains challenging, as higher values may degrade reconstruction fidelity. 

To more directly minimize statistical dependencies, FactorVAE \cite{kim2018disentangling} employs an adversarial learning strategy. It introduces a discriminator to distinguish between latent codes sampled from the aggregated posterior distribution and those from an independent prior distribution, thereby explicitly penalizing the Total Correlation (TC) of latent variables. This approach directly optimizes the aggregated posterior to approximate an independent distribution.

Building upon this, $\beta$-TCVAE \cite{chen2018isolating} decomposes the KL divergence term in the VAE objective into three components: index-code mutual information, total correlation (TC), and dimension-wise KL divergence. By assigning distinct weights to these components, $\beta$-TCVAE achieves finer-grained control over latent dependencies. Its central innovation lies in explicitly penalizing TC to reduce interdimensional correlations.
\paragraph{GAN-based Methods.} Within the domain of GAN-based methods, InfoGAN \cite{chen2016infogan} a significant contribution, giving rise to a series of InfoGAN-based variants, including InfoGAN-CR\cite{lin2020infogan} and PS-SC GAN\cite{zhu2021and}. The objective of InfoGAN is to learn interpretable latent representations by maximizing the mutual information between a subset of the latent variables and the output of the generator. Although its primary objective is not the direct minimization of statistical dependencies across all latent variables, InfoGAN indirectly facilitates a reduction in statistical dependencies between specific latent variables encoding semantic features and the remaining latent variables. This effect contributes to the emergence of disentangled representations. 

InfoGAN-CR, a variant of InfoGAN with a contrastive regularizer, generates multiple images by fixing one dimension, denoted $c_i$, of the latent representation, while randomly sampling the other dimensions, denoted $c_j(i \neq j)$. A classifier is then trained to identify which latent dimension was fixed based on the generated images. The contrastive regularizer promotes differentiation among latent representation dimensions, thereby supporting disentanglement.

\begin{figure*}[htbp] 
\centering 
\includegraphics[width=\textwidth]{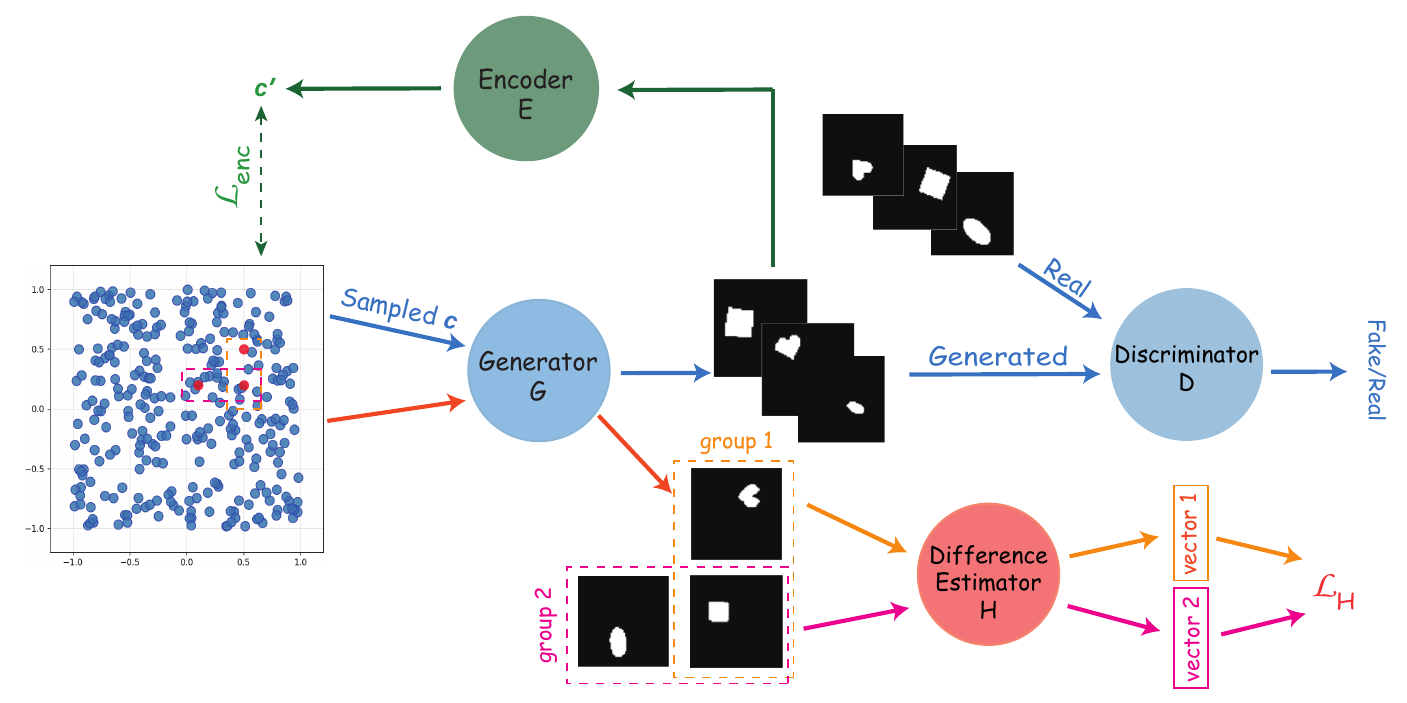} 
\caption{ The framework of the proposed DiD. The blue points denote $\mathbf{c}$ sampled from the uniform distribution, the red points represent samples obtained from two orthogonal axes.} 
\label{fig::model} 
\end{figure*}

\section{Methodology}
\subsection{Formulation}
Many existing disentangled representation learning methods often indirectly achieve disentanglement by penalizing the statistical dependencies among latent variables. However, reducing statistical dependencies does not fully guarantee improved disentanglement capability. Therefore, this paper aims to design a learning paradigm that directly promotes the enhancement of disentanglement capability, starting from the semantic differences between different latent factors.

We assume that the semantic differences between distinct latent factors should be reflected in the variations of a series of samples across their attributes, and these variation patterns should exhibit human-interpretable regularity. In other words, the semantic variations of latent factors arise from the differences and changes among samples, while these differences are not present in individual samples. Since the semantic differences of factors originate from sample differences, the model should learn the semantic differences of factors from these sample differences, thereby directly promoting the improvement of the model's disentanglement capability.

To learn sample differences, we first need to map the samples from the dataset into a latent space that conforms to the dataset distribution. At this stage, the latent space is relatively disordered, and individual dimensions do not effectively represent factors. Subsequently, we employ a pairwise contrastive disentanglement prior to organize this disordered latent space, gradually enabling the latent space to exhibit characteristics of mutually independent and orderly arranged factors. Specifically, the pairwise contrastive method posits that for sample pairs dominated by variations in the same latent factor, the vector difference in the representation space should be significantly smaller than that of sample pairs driven by variations in different latent factors.  Essentially, variations in samples caused by different latent factors should be mapped to larger distances in the representation space, while variations caused by the same factor should be mapped to closer distances. Thus, by encoding and amplifying the distances in the representation space corresponding to sample variations driven by different factors, we can encourage the alignment of samples with similar variation patterns along the same dimension of the representation space, thereby achieving effective disentanglement of latent factors.

\subsection{Framework of the DiD}

Based on the aforementioned disentanglement principles, this study proposes the Disentanglement in Difference (DiD) model. This model primarily consists of three core modules: a Sample Generation Paradigm responsible for learning the real data distribution and generating high-quality samples; a Difference Encoder, designed to encode the differences between sample pairs caused by distinct latent factors and map these differences to a representation space, thereby capturing the influence of latent factors; and a Sample Encoder, used to map samples to the latent space and obtain their representations within this space.

As shown in Figure \ref{fig::model}, the workflow of DiD is as follows: As the sample generation component for disentangled representation learning, a vector $\mathbf{c}$ is first sampled from an $n$-dimensional latent space $S$ as input to a generator $G$, generating a sample $\mathbf{x}$. The generated sample $\mathbf{x}$ simultaneously serves as input to both a discriminator $D$ and the Sample Encoder for their respective training. This ensures the model's ability to generate high-quality samples while also obtaining latent space representations of the input samples. To achieve factor disentanglement, we select two latent representations from the latent space $S$, each controlled by specific dimensions, and input these two representations separately into the generator $G$, thereby obtaining two sets of sample pairs. Specifically, it is assumed that within each sample pair, the two samples differ only in one latent space dimension, while remaining consistent across other factors. These two sets of sample pairs are then fed into the Difference Encoder, yielding two vectors $\mathbf{v}_1$ and $\mathbf{v}_2$. These vectors respectively represent the variations between the corresponding sample pairs caused by specific latent factors. Consequently, the distance between $\mathbf{v}_1$ and $\mathbf{v}_2$ should be maximized. However, as the sample pairs are selected from dimensions within a disordered latent space, $\mathbf{v}_1$ and $\mathbf{v}_2$ are not initially maximized. By maximizing the distance between vectors $\mathbf{v}_1$ and $\mathbf{v}_2$, we encourage the generator $G$ to generate disentangled sample variations along independent latent factors within the latent space $S$.

\subsection{Samples Generation Paradigm}

The sample generation paradigm employs a general sample generation network, such as DCGAN\cite{radford2015unsupervised}, WGAN\cite{arjovsky2017wasserstein}, etc. In this paper, WGAN-GP\cite{gulrajani2017improved} is utilized as the sample generation paradigm to maintain high-quality generation performance.

To establish a correspondence between each dimension of the latent space and a single factor, it is necessary to construct a low-dimensional latent space that effectively represents the samples of the dataset and supports free and continuous exploration within this space. We need to rationally design the distribution of the latent space. Specifically, we desire the latent space to possess finite boundaries to prevent the model from over-exploration or getting trapped in irrelevant regions. Simultaneously, to encourage the model to fully utilize the capacity of the latent space, learn complete information about the data distribution, and mitigate the phenomenon of latent representation aggregation, we choose a bounded prior distribution that follows a uniform distribution $c \sim U[-1, 1]^n$ as the prior constraint for the latent space. To effectively map samples from the real dataset into this latent space, sample points from this space are used as input to the sample generation paradigm. The generator and discriminator are optimized by minimizing the Earth-Mover distance (also known as the Wasserstein distance) between the generated distribution and the real distribution, ensuring that the latent space can effectively fit the real distribution.
The WGAN-GP loss is given by:
\begin{equation}
\mathcal{L}_{G} = - \mathbb{E}_{c \sim p(c)} [D(G(c))]
\end{equation}
\begin{equation}
\begin{split}
    \mathcal{L}_{D} = &\mathbb{E}_{x \sim p_{\text{data}}(x)} [D(x)] - \mathbb{E}_{c \sim p(c)} [D(G(c))] \\&+ \lambda \mathbb{E}_{\hat{x} \sim p_{\hat{x}}} [(||\nabla_{\hat{x}} D(\hat{x})||_2 - 1)^2]
\end{split}
\end{equation}

\subsection{Difference Encoder}

To enable the model to learn differentiated representations that effectively capture variations in distinct latent factors, we introduce a Difference Encoder, denoted as $H$. This network architecture can be realized using multi-layer Convolutional Neural Networks (CNNs) or Multi-Layer Perceptrons (MLPs).  The objective of the Difference Encoder is to maximize the distance in the representation space between generated sample pairs resulting from variations in different latent factors. 

Specifically, we first sample a latent code, $c$, from an $n$-dimensional latent space, $S$. We then select two dimensions, denoted $a_1$ and $a_2$, from $S$ centered at $c$. These dimensions correspond to two mutually independent latent factors that we aim to disentangle. To represent variations controlled by these factors, we perturb $c$ along these dimensions, obtaining new latent codes $c'$ and $c''$. For instance, $c'$ is obtained by moving a short distance from $c$ along the $a_1$ direction, while $c''$ is obtained by moving a short distance from $c$ along the $a_2$ direction. Subsequently, we input $c$, $c'$, and $c''$ to the generator, $G$, to obtain the corresponding generated samples: $x = G(c)$, $x' = G(c')$, and $x'' = G(c'')$.  To learn the semantic difference, two sets of sample pairs are constructed: $g_1 = [x, x']$ and $g_2 = [x, x'']$. The sample pair $(x, x')$ in $g_1$ represents a variation in latent factor 1 (corresponding to $a_1$), while the sample pair $(x, x'')$ in $g_2$ represents a variation in latent factor 2 (corresponding to $a_2$). Next, the sample pairs $g_1$ and $g_2$ are fed into the Difference Encoder, $H$, yielding the corresponding difference encoding vectors, $v_1 = H(g_1)$ and $v_2 = H(g_2)$. Our objective is to maximize the distance, $d(v_1, v_2)$, between $v_1$ and $v_2$, where the Euclidean distance is employed. 

This procedure aims to ensure that when generated sample differences are controlled by different latent factors, their corresponding encoding vectors in the difference representation space exhibit greater separation. In effect, the Difference Encoder effectively captures and distinguishes variations induced by different latent factors.

The contrastive loss is given by:
\begin{equation}
\mathcal{L}_{H} = - ||v_1 - v_2||_2 = - ||H(g_1) - H(g_2)||_2
\end{equation}

\subsection{Samples Encoder}

Since disentangled representation models typically need to possess the capability to obtain effective latent space representations of samples, we introduce a Sample Encoder $E$ to achieve this. This encoder takes the samples $G({c})$ generated by the sample generation paradigm as input and aims to output the encoded representation ${c}'_{enc}$ of this sample in the latent space. Our goal is to make the encoder's output ${c}'_{enc}$ as close as possible to the original latent code ${c}$ used to generate the sample. To achieve this objective, we employ a method of minimizing the difference between the encoder's output and the generator's input to train the encoder. Specifically, we optimize the encoder's parameters by minimizing the Mean Squared Error (MSE) loss function $L_{enc} = ||{c} - {c}'_{enc}||^2$. Through iterative optimization, we expect the encoder $E$ to learn a mapping from the generated sample space to the latent space, such that for a given generated sample, the encoder can accurately map it back to its corresponding location in the latent space.
The encoder reconstruction loss is given by:
\begin{equation}
\mathcal{L}_{enc} = ||c - c'_{enc}||_2^2 = ||c - E(G(c))||_2^2
\end{equation}

The total loss function when training the generator is:
\begin{equation}
\mathcal{L} = \mathcal{L}_{G} + \mathcal{L}_{H} + \mathcal{L}_{enc}
\end{equation}
The total loss function when training the discriminator is:
\begin{equation}
\mathcal{L} = \mathcal{L}_{D}
\end{equation}

\begin{table*}[t]
\centering
\caption{Disentanglement Performance Comparison on dSprites and 3DShapes Datasets}
\label{tab:results}
\begin{tabularx}{\textwidth}{@{}>{\centering\arraybackslash}X >{\centering\arraybackslash}X >{\centering\arraybackslash}X >{\centering\arraybackslash}X >{\centering\arraybackslash}X@{}}
\toprule
\multirow{2}{*}{Dataset} & \multirow{2}{*}{Model} & \multicolumn{3}{c}{Metrics} \\
\cmidrule(lr){3-5}
 & & \textbf{MIG} & \textbf{DCI-D} & \textbf{SAP} \\
\midrule
\multirow{6}{*}{dSprites} 
& $\beta$-VAE (4.0)    & 0.24 ± 0.04   & 0.42 ± 0.11   & 0.10 ± 0.03   \\
& $\beta$-TCVAE (6.0)  & 0.32 ± 0.06   & 0.49 ± 0.10   & 0.08 ± 0.02   \\
& FactorVAE            & 0.38 ± 0.06   & 0.55 ± 0.05   & \textbf{0.19 ± 0.03} \\
& DynamicVAE           & 0.50 ± 0.06   & 0.55 ± 0.04 & 0.13 ± 0.03   \\
& InfoGAN-CR           & 0.25 ± 0.02   & 0.36 ± 0.01   & 0.12 ± 0.01   \\
& DiD                 & \textbf{0.66 ± 0.03} & \textbf{0.58 ± 0.06}   & 0.17 ± 0.02   \\
\cmidrule(lr){1-5}
\multirow{6}{*}{3DShapes}
& $\beta$-VAE (4.0)    & 0.47 ± 0.01   & 0.32 ± 0.03   & 0.12 ± 0.03   \\
& $\beta$-TCVAE (6.0)  & 0.49 ± 0.02   & 0.37 ± 0.04   & 0.15 ± 0.02   \\
& FactorVAE            & 0.39 ± 0.04   & 0.38 ± 0.03   & 0.12 ± 0.04   \\
& DynamicVAE           & 0.58 ± 0.06   & 0.42 ± 0.02   & 0.11 ± 0.02   \\
& InfoGAN-CR           & 0.51 ± 0.01   & 0.37 ± 0.01   & 0.14 ± 0.03   \\
& DiD                  & \textbf{0.63 ± 0.02} & \textbf{0.49 ± 0.01} & \textbf{0.24 ± 0.06} \\
\bottomrule
\end{tabularx}
\end{table*}

\section{EXPERIMENTS}

\subsection{Datasets}
To systematically evaluate the effectiveness of the proposed Differentiable Information Disentanglement (DiD) model, we selected two widely adopted datasets in the disentangled representation learning domain: dSprites\cite{dsprites17} and 3DShapes\cite{3dshapes18}. Both datasets are known for their controlled generative processes and clearly defined latent factors, providing an ideal platform for the quantitative assessment of disentangling performance. 

The dSprites dataset\cite{dsprites17} consists of 737,280 binarized images with a size of $64 \times 64$ pixels and single-channel grayscale. The generative process of this dataset is controlled by five disentangled latent factors: shape (three categories: square, ellipse, heart), size (six discrete scales), rotation angle (40 evenly distributed angles), and translations in both the horizontal (x-axis) and vertical (y-axis) directions (each with 32 discrete positions). These factors are designed to change independently during the generation process, ensuring clarity and controllability of the dataset's latent structure.

The 3DShapes dataset\cite{3dshapes18} contains 480,000 RGB color images with a resolution of $64 \times 64$ pixels. Its generative process is driven by six latent factors: floor color (10 types), wall color (10 types), object color (10 types), object size (8 types), object shape (4 types), and object angle (15 angles).

\subsection{Evaluation Metrics}
While a standard metric for measuring disentanglement is lacking\cite{zhou2020evaluating,ridgeway2018learning}, numerous approaches exist to evaluate it based on intervention, predictor, and information\cite{carbonneau2022metrics}.
To quantitatively evaluate the disentangling performance of the model and make fair comparisons with existing methods, we adopted three widely recognized and complementary disentangling evaluation metrics: Mutual Information Gap (MIG) \cite{chen2018isolating}, Disentanglement, Completeness, and Informativeness (DCI) \cite{eastwood2018DCI}, and Separated Attribute Predictability (SAP) \cite{kumar2017SAP}.

The MIG metric evaluates the degree of disentanglement by calculating the mutual information between each latent dimension and its most relevant attribute, then measuring the gap between the highest and second-highest mutual information. A higher MIG value indicates a stronger correlation between the learned representation and the latent factors, reflecting better disentangling performance.

The DCI framework evaluates the quality of representations from three dimensions: disentanglement, completeness, and informativeness. In this study, we focus primarily on the disentanglement metric (DCI-D). DCI-D is calculated by training linear classifiers that predict the values of latent factors based on individual latent dimensions. The disentanglement score is computed based on the accuracy of these classifiers. Higher accuracy implies that the model can effectively disentangle different factors into separate latent dimensions.

The SAP metric directly assesses whether a single latent dimension can accurately predict a specific attribute in the data. SAP is calculated by identifying the attribute that can be best predicted by each latent dimension and computing the prediction accuracy. A higher SAP value indicates that the model's latent dimensions correspond more strongly to the data attributes, reflecting better disentanglement.

\subsection{Implementation Details}
To ensure the reproducibility and fairness of the experiments, we describe the implementation details of the DiD model as follows:

The DiD model adopts a modular network design, including a generator, discriminator, encoder, and difference encoder. The generator uses a deconvolutional neural network (DCNN) structure\cite{zeiler2014deconv}, responsible for generating images from the latent space. Both the discriminator and encoder use convolutional neural networks (CNNs)\cite{lecun1998cnn}, which distinguish real images from generated ones and map input images to the latent space, respectively. The difference encoder employs a multi-layer perceptron (MLP) to estimate the difference information in the latent representations. All network layers use LeakyReLU activation functions to enhance the model’s non-linear fitting capacity and training stability. The parameters of the generator, encoder, and difference encoder are optimized using the RMSProp optimizer with a learning rate of $1 \times 10^{-4}$. The discriminator is optimized using the Adam optimizer with a learning rate of $1 \times 10^{-4}$, and momentum parameters $\beta_1 = 0.5$ and $\beta_2 = 0.999$. All models are trained using batch gradient descent with a batch size of 128.

\subsection{Quantitative Analysis}
To quantify the disentangling performance of the DiD model and compare it with current mainstream disentangling methods, we conducted comprehensive experimental evaluations on the dSprites and 3DShapes datasets. The DiD model was compared with a series of representative baseline models, including $\beta$-VAE \cite{higgins2017beta}, $\beta$-TCVAE \cite{chen2018isolating}, FactorVAE \cite{van2019disentangled}, Dynamic-VAE \cite{shao2020dynamicvae}, and InfoGAN-CR \cite{lin2020infogan}. All models were trained for 500,000 steps on both datasets, and the disentangling performance was quantitatively assessed using the MIG, DCI-D, and SAP metrics.

To ensure the fairness of the comparison, we conducted a detailed hyperparameter tuning for all baseline models and report the best hyperparameter configurations that achieved the highest disentangling performance on the validation set. Specifically, for $\beta$-VAE, we selected $\beta = 4$; for $\beta$-TCVAE, we selected $\beta = 6$; for FactorVAE, we selected $\gamma = 4$; and for Dynamic-VAE, we selected $K_i=0.001,K_p=0.01$.

Table \ref{tab:results} summarizes the disentangling metric scores of the DiD model and each baseline model on the dSprites and 3DShapes datasets. The experimental results clearly demonstrate that the DiD model consistently achieves significantly higher SAP, MIG, and DCI-D scores compared to all baseline methods. Notably, the DiD model exhibits a particularly strong advantage in the MIG metric, providing compelling evidence that the DiD model effectively reduces mutual information between representation dimensions. Even without explicitly penalizing the statistical dependence between latent variables, the model achieves effective disentangling of the latent space dimensions, learning better disentangled representations.

\subsection{Total Correlation and Disentangling Metrics}
As mentioned earlier, existing methods encourage disentangling by penalizing the statistical dependence between latent variables (e.g., total correlation). However, these two aspects are not strictly equivalent. To verify this hypothesis, we conducted a comparative analysis of the dynamic relationship between total correlation (TC) and disentangling metrics (MIG) during the training process of $\beta$-VAE, $\beta$-TCVAE, and FactorVAE. The experiment used the dSprites dataset, with hyperparameters fixed at $\beta=4$ and batch size = 128 to control variable effects.

\begin{figure}
    \centering
    \includegraphics[width=0.8\linewidth]{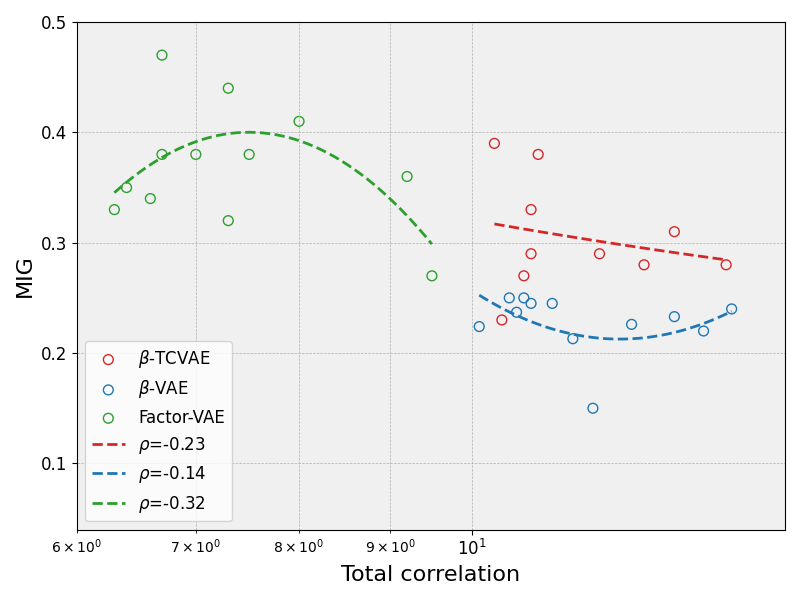}
    \caption{Relationship between Total Correlation (TC) and Mutual Information Gap (MIG). The scatter plots illustrate that a lower Total Correlation does not consistently guarantee improved disentanglement performance.}
    \label{fig2}
\end{figure}

As shown in Figure \ref{fig2}, while training effectively reduced total correlation among latent variables in all three models, the Mutual Information Gap (MIG) metric displayed considerable volatility and failed to exhibit a corresponding consistent improvement. In fact, a declining trend in MIG was sometimes evident. This indicates that the reduction of statistical dependencies and the enhancement of disentanglement are not directly proportional. Consequently, directly penalizing statistical dependencies alone may not be a robust method for achieving reliable disentanglement.

\subsection{Ablation Study on the Difference Estimator}
This section presents an investigation into the core functionality of the Difference Estimator within our proposed model framework. Our analysis revolves around two key directions: firstly, exploring the indispensable role of the Difference Estimator for disentangled representation learning, and secondly, examining the quantitative effect of manipulating the number of comparable dimensions within the Difference Estimator on model disentanglement capabilities. 

\begin{figure}
    \centering
    \includegraphics[width=0.8\linewidth]{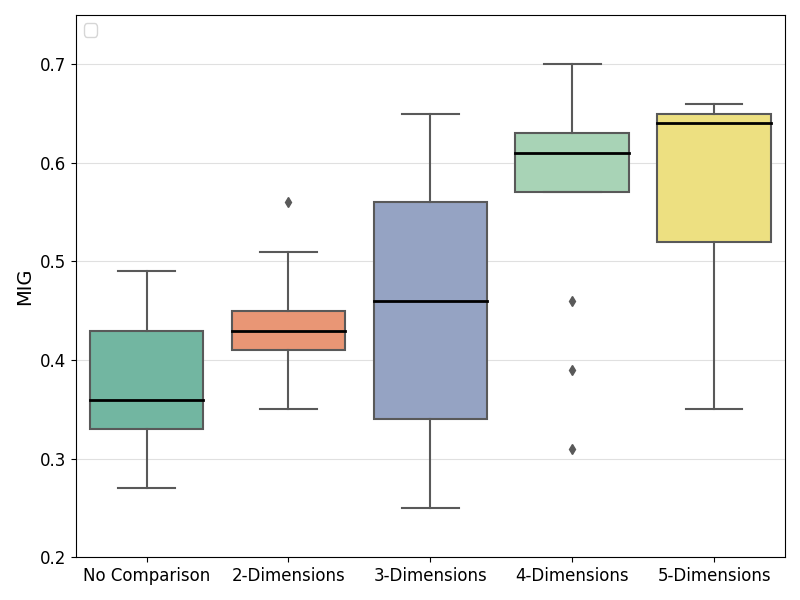}
    \caption{In this ablation study, model performance was investigated by varying the number of dimensions used for difference comparison. The x-axis indicates the count of dimensions selected from the total number of dimensions for comparison.}
    \label{fig3}
\end{figure}

As shown in Figure \ref{fig3}, the ablation of the Difference Estimator module (labeled 'No Comparison') results in the functional devolution of the model, resembling a Wasserstein GAN with Gradient Penalty (WGAN-GP). In this scenario, the results clearly showed that the model struggled to disentangle the factors, reflected in significantly lower disentanglement scores. However, progressively increasing the number of dimensions for comparison within the Difference Estimator reveals a distinct enhancement in disentanglement performance. Quantitative data demonstrate that this augmentation effectively extends the model's capacity for disentangling representations to a higher-dimensional latent space, which strongly indicates that our difference comparison mechanism facilitates the effective learning of disentangled representations.

\section{Conclusion}
This paper introduces a novel disentanglement representation learning approach that learns by encoding differences between latent factors. By encoding semantic differences between factors, this approach encourages the model to map disparate factors to more distant regions in the latent space, aligning with the intuition that semantically distinct factors should exhibit greater separation. On simple datasets comprised of explicit factors, we demonstrate our method's effectiveness at achieving strong disentanglement performance. Furthermore, an ablation study confirms that explicitly learning semantic differences between factors significantly enhances the model's disentanglement capabilities.
% Acknowledgements should only appear in the accepted version.
\section*{Acknowledgements}
This work was supported by National Natural Science Foundation of China under Grant No. 61872419, No. 62072213. Shandong Provincial Natural Science Foundation No. ZR2022JQ30, No. ZR2022ZD01, No. ZR2023LZH015. Taishan Scholars Program of Shandong Province, China, under Grant No. tsqn201812077. “New 20 Rules for University” Program of Jinan City under Grant No. 2021GXRC077. Key Research Project of Quancheng Laboratory, China under Grant No. QCLZD202303. Research Project of Provincial Laboratory of Shandong, China under Grant No. SYS202201.
% In the unusual situation where you want a paper to appear in the
% references without citing it in the main text, use \nocite
\nocite{langley00}

\bibliography{example_paper}

\begin{thebibliography}{36}
\providecommand{\natexlab}[1]{#1}
\providecommand{\url}[1]{\texttt{#1}}
\expandafter\ifx\csname urlstyle\endcsname\relax
  \providecommand{\doi}[1]{doi: #1}\else
  \providecommand{\doi}{doi: \begingroup \urlstyle{rm}\Url}\fi

\bibitem[Arjovsky et~al.(2017)Arjovsky, Chintala, and Bottou]{arjovsky2017wasserstein}
Arjovsky, M., Chintala, S., and Bottou, L.
\newblock Wasserstein generative adversarial networks.
\newblock In \emph{International conference on machine learning}, pp.\  214--223. PMLR, 2017.

\bibitem[Bengio et~al.(2013)Bengio, Courville, and Vincent]{bengio2013representation}
Bengio, Y., Courville, A., and Vincent, P.
\newblock Representation learning: A review and new perspectives.
\newblock \emph{IEEE transactions on pattern analysis and machine intelligence}, 35\penalty0 (8):\penalty0 1798--1828, 2013.

\bibitem[Bottou et~al.(2007)Bottou, Chapelle, DeCoste, and Weston]{bottou2007scalling}
Bottou, L., Chapelle, O., DeCoste, D., and Weston, J.
\newblock Scaling learning algorithms toward ai.
\newblock 2007.

\bibitem[Burgess \& Kim(2018)Burgess and Kim]{3dshapes18}
Burgess, C. and Kim, H.
\newblock 3d shapes dataset.
\newblock https://github.com/deepmind/3dshapes-dataset/, 2018.

\bibitem[Burgess et~al.(2018)Burgess, Higgins, Pal, Matthey, Watters, Desjardins, and Lerchner]{burgess2018understanding}
Burgess, C.~P., Higgins, I., Pal, A., Matthey, L., Watters, N., Desjardins, G., and Lerchner, A.
\newblock Understanding disentangling in $\beta$-vae.
\newblock \emph{arXiv preprint arXiv:1804.03599}, 2018.

\bibitem[Carbonneau et~al.(2022)Carbonneau, Zaidi, Boilard, and Gagnon]{carbonneau2022metrics}
Carbonneau, M.-A., Zaidi, J., Boilard, J., and Gagnon, G.
\newblock Measuring disentanglement: A review of metrics.
\newblock \emph{IEEE transactions on neural networks and learning systems}, 2022.

\bibitem[Chen et~al.(2018)Chen, Li, Grosse, and Duvenaud]{chen2018isolating}
Chen, R.~T., Li, X., Grosse, R.~B., and Duvenaud, D.~K.
\newblock Isolating sources of disentanglement in variational autoencoders.
\newblock \emph{Advances in neural information processing systems}, 31, 2018.

\bibitem[Chen et~al.(2016)Chen, Duan, Houthooft, Schulman, Sutskever, and Abbeel]{chen2016infogan}
Chen, X., Duan, Y., Houthooft, R., Schulman, J., Sutskever, I., and Abbeel, P.
\newblock Infogan: Interpretable representation learning by information maximizing generative adversarial nets.
\newblock \emph{Advances in neural information processing systems}, 29, 2016.

\bibitem[Eastwood \& Williams(2018)Eastwood and Williams]{eastwood2018DCI}
Eastwood, C. and Williams, C.~K.
\newblock A framework for the quantitative evaluation of disentangled representations.
\newblock In \emph{6th International Conference on Learning Representations}, 2018.

\bibitem[Gabbay \& Hoshen(2019)Gabbay and Hoshen]{gabbay2019demystifying}
Gabbay, A. and Hoshen, Y.
\newblock Demystifying inter-class disentanglement.
\newblock \emph{arXiv preprint arXiv:1906.11796}, 2019.

\bibitem[Gabbay \& Hoshen(2021)Gabbay and Hoshen]{gabbay2021scaling}
Gabbay, A. and Hoshen, Y.
\newblock Scaling-up disentanglement for image translation.
\newblock In \emph{Proceedings of the IEEE/CVF International Conference on Computer Vision}, pp.\  6783--6792, 2021.

\bibitem[Gulrajani et~al.(2017)Gulrajani, Ahmed, Arjovsky, Dumoulin, and Courville]{gulrajani2017improved}
Gulrajani, I., Ahmed, F., Arjovsky, M., Dumoulin, V., and Courville, A.~C.
\newblock Improved training of wasserstein gans.
\newblock \emph{Advances in neural information processing systems}, 30, 2017.

\bibitem[Higgins et~al.(2017)Higgins, Matthey, Pal, Burgess, Glorot, Botvinick, Mohamed, and Lerchner]{higgins2017beta}
Higgins, I., Matthey, L., Pal, A., Burgess, C.~P., Glorot, X., Botvinick, M.~M., Mohamed, S., and Lerchner, A.
\newblock beta-vae: Learning basic visual concepts with a constrained variational framework.
\newblock \emph{ICLR (Poster)}, 3, 2017.

\bibitem[Higgins et~al.(2018)Higgins, Amos, Pfau, Racaniere, Matthey, Rezende, and Lerchner]{higgins2018towards}
Higgins, I., Amos, D., Pfau, D., Racaniere, S., Matthey, L., Rezende, D., and Lerchner, A.
\newblock Towards a definition of disentangled representations.
\newblock \emph{arXiv preprint arXiv:1812.02230}, 2018.

\bibitem[Kim \& Mnih(2018)Kim and Mnih]{kim2018disentangling}
Kim, H. and Mnih, A.
\newblock Disentangling by factorising.
\newblock In \emph{International conference on machine learning}, pp.\  2649--2658. PMLR, 2018.

\bibitem[Kingma(2013)]{kingma2013auto}
Kingma, D.~P.
\newblock Auto-encoding variational bayes.
\newblock \emph{arXiv preprint arXiv:1312.6114}, 2013.

\bibitem[Kumar et~al.(2017{\natexlab{a}})Kumar, Sattigeri, and Balakrishnan]{kumar2017SAP}
Kumar, A., Sattigeri, P., and Balakrishnan, A.
\newblock Variational inference of disentangled latent concepts from unlabeled observations.
\newblock \emph{arXiv preprint arXiv:1711.00848}, 2017{\natexlab{a}}.

\bibitem[Kumar et~al.(2017{\natexlab{b}})Kumar, Sattigeri, and Balakrishnan]{kumar2017variational}
Kumar, A., Sattigeri, P., and Balakrishnan, A.
\newblock Variational inference of disentangled latent concepts from unlabeled observations.
\newblock \emph{arXiv preprint arXiv:1711.00848}, 2017{\natexlab{b}}.

\bibitem[Lake et~al.(2017)Lake, Ullman, Tenenbaum, and Gershman]{lake2017building}
Lake, B.~M., Ullman, T.~D., Tenenbaum, J.~B., and Gershman, S.~J.
\newblock Building machines that learn and think like people.
\newblock \emph{Behavioral and brain sciences}, 40:\penalty0 e253, 2017.

\bibitem[LeCun et~al.(1998)LeCun, Bottou, Bengio, and Haffner]{lecun1998cnn}
LeCun, Y., Bottou, L., Bengio, Y., and Haffner, P.
\newblock Gradient-based learning applied to document recognition.
\newblock \emph{Proceedings of the IEEE}, 86\penalty0 (11):\penalty0 2278--2324, 1998.

\bibitem[Lin et~al.(2020)Lin, Thekumparampil, Fanti, and Oh]{lin2020infogan}
Lin, Z., Thekumparampil, K., Fanti, G., and Oh, S.
\newblock Infogan-cr and modelcentrality: Self-supervised model training and selection for disentangling gans.
\newblock In \emph{international conference on machine learning}, pp.\  6127--6139. PMLR, 2020.

\bibitem[Locatello et~al.(2019{\natexlab{a}})Locatello, Bauer, Lucic, Raetsch, Gelly, Sch{\"o}lkopf, and Bachem]{locatello2019challenging}
Locatello, F., Bauer, S., Lucic, M., Raetsch, G., Gelly, S., Sch{\"o}lkopf, B., and Bachem, O.
\newblock Challenging common assumptions in the unsupervised learning of disentangled representations.
\newblock In \emph{international conference on machine learning}, pp.\  4114--4124. PMLR, 2019{\natexlab{a}}.

\bibitem[Locatello et~al.(2019{\natexlab{b}})Locatello, Tschannen, Bauer, R{\"a}tsch, Sch{\"o}lkopf, and Bachem]{locatello2019disentangling}
Locatello, F., Tschannen, M., Bauer, S., R{\"a}tsch, G., Sch{\"o}lkopf, B., and Bachem, O.
\newblock Disentangling factors of variation using few labels.
\newblock \emph{arXiv preprint arXiv:1905.01258}, 2019{\natexlab{b}}.

\bibitem[Matthey et~al.(2017)Matthey, Higgins, Hassabis, and Lerchner]{dsprites17}
Matthey, L., Higgins, I., Hassabis, D., and Lerchner, A.
\newblock dsprites: Disentanglement testing sprites dataset.
\newblock https://github.com/deepmind/dsprites-dataset/, 2017.

\bibitem[Peters et~al.(2017)Peters, Janzing, and Sch{\"o}lkopf]{peters2017elements}
Peters, J., Janzing, D., and Sch{\"o}lkopf, B.
\newblock \emph{Elements of causal inference: foundations and learning algorithms}.
\newblock The MIT Press, 2017.

\bibitem[Radford(2015)]{radford2015unsupervised}
Radford, A.
\newblock Unsupervised representation learning with deep convolutional generative adversarial networks.
\newblock \emph{arXiv preprint arXiv:1511.06434}, 2015.

\bibitem[Ridgeway \& Mozer(2018)Ridgeway and Mozer]{ridgeway2018learning}
Ridgeway, K. and Mozer, M.~C.
\newblock Learning deep disentangled embeddings with the f-statistic loss.
\newblock \emph{Advances in neural information processing systems}, 31, 2018.

\bibitem[Schmidhuber et~al.(1996)Schmidhuber, Eldracher, and Foltin]{schmidhuber1996semilinear}
Schmidhuber, J., Eldracher, M., and Foltin, B.
\newblock Semilinear predictability minimization produces well-known feature detectors.
\newblock \emph{Neural Computation}, 8\penalty0 (4):\penalty0 773--786, 1996.

\bibitem[Shao et~al.(2020)Shao, Lin, Yang, Yao, Zhao, and Abdelzaher]{shao2020dynamicvae}
Shao, H., Lin, H., Yang, Q., Yao, S., Zhao, H., and Abdelzaher, T.
\newblock Dynamicvae: Decoupling reconstruction error and disentangled representation learning.
\newblock \emph{arXiv preprint arXiv:2009.06795}, 2020.

\bibitem[Tschannen et~al.(2018)Tschannen, Bachem, and Lucic]{tschannen2018recent}
Tschannen, M., Bachem, O., and Lucic, M.
\newblock Recent advances in autoencoder-based representation learning.
\newblock \emph{arXiv preprint arXiv:1812.05069}, 2018.

\bibitem[Van~Steenkiste et~al.(2019)Van~Steenkiste, Locatello, Schmidhuber, and Bachem]{van2019disentangled}
Van~Steenkiste, S., Locatello, F., Schmidhuber, J., and Bachem, O.
\newblock Are disentangled representations helpful for abstract visual reasoning?
\newblock \emph{Advances in neural information processing systems}, 32, 2019.

\bibitem[Watanabe(1960)]{watanabe1960information}
Watanabe, S.
\newblock Information theoretical analysis of multivariate correlation.
\newblock \emph{IBM Journal of research and development}, 4\penalty0 (1):\penalty0 66--82, 1960.

\bibitem[Zeiler \& Fergus(2014)Zeiler and Fergus]{zeiler2014deconv}
Zeiler, M.~D. and Fergus, R.
\newblock Visualizing and understanding convolutional networks.
\newblock In \emph{Computer Vision--ECCV 2014: 13th European Conference, Zurich, Switzerland, September 6-12, 2014, Proceedings, Part I 13}, pp.\  818--833. Springer, 2014.

\bibitem[Zhou et~al.(2020)Zhou, Zelikman, Lu, Ng, Carlsson, and Ermon]{zhou2020evaluating}
Zhou, S., Zelikman, E., Lu, F., Ng, A.~Y., Carlsson, G., and Ermon, S.
\newblock Evaluating the disentanglement of deep generative models through manifold topology.
\newblock \emph{arXiv preprint arXiv:2006.03680}, 2020.

\bibitem[Zhu et~al.(2018)Zhu, Zhang, Zhang, Wu, Torralba, Tenenbaum, and Freeman]{zhu2018visual}
Zhu, J.-Y., Zhang, Z., Zhang, C., Wu, J., Torralba, A., Tenenbaum, J., and Freeman, B.
\newblock Visual object networks: Image generation with disentangled 3d representations.
\newblock \emph{Advances in neural information processing systems}, 31, 2018.

\bibitem[Zhu et~al.(2021)Zhu, Xu, and Tao]{zhu2021and}
Zhu, X., Xu, C., and Tao, D.
\newblock Where and what? examining interpretable disentangled representations.
\newblock In \emph{Proceedings of the IEEE/CVF Conference on Computer Vision and Pattern Recognition}, pp.\  5861--5870, 2021.

\end{thebibliography}
\bibliographystyle{icml2025}

%%%%%%%%%%%%%%%%%%%%%%%%%%%%%%%%%%%%%%%%%%%%%%%%%%%%%%%%%%%%%%%%%%%%%%%%%%%%%%%
%%%%%%%%%%%%%%%%%%%%%%%%%%%%%%%%%%%%%%%%%%%%%%%%%%%%%%%%%%%%%%%%%%%%%%%%%%%%%%%
% APPENDIX
%%%%%%%%%%%%%%%%%%%%%%%%%%%%%%%%%%%%%%%%%%%%%%%%%%%%%%%%%%%%%%%%%%%%%%%%%%%%%%%
%%%%%%%%%%%%%%%%%%%%%%%%%%%%%%%%%%%%%%%%%%%%%%%%%%%%%%%%%%%%%%%%%%%%%%%%%%%%%%%
\newpage
\appendix
\onecolumn
% \section{You \emph{can} have an appendix here.}

% You can have as much text here as you want. The main body must be at most $8$ pages long.
% For the final version, one more page can be added.
% If you want, you can use an appendix like this one.  

% The $\mathtt{\backslash onecolumn}$ command above can be kept in place if you prefer a one-column appendix, or can be removed if you prefer a two-column appendix.  Apart from this possible change, the style (font size, spacing, margins, page numbering, etc.) should be kept the same as the main body.
%%%%%%%%%%%%%%%%%%%%%%%%%%%%%%%%%%%%%%%%%%%%%%%%%%%%%%%%%%%%%%%%%%%%%%%%%%%%%%%
%%%%%%%%%%%%%%%%%%%%%%%%%%%%%%%%%%%%%%%%%%%%%%%%%%%%%%%%%%%%%%%%%%%%%%%%%%%%%%%

\end{document}